\documentclass[11pt]{article}

\usepackage[margin=1in]{geometry}
\usepackage[T1]{fontenc}
\usepackage{lmodern}
\usepackage{amsmath,amssymb}
\usepackage{graphicx}
\usepackage{booktabs}
\usepackage{array}
\usepackage{multirow}
\usepackage{xcolor}
\usepackage{xspace}
\xspaceaddexceptions{\, \( \%}%
\usepackage[numbers,sort&compress,super]{natbib}
\usepackage{caption}
\usepackage[hidelinks]{hyperref}
\usepackage{xurl}%

\newcommand{\pending}[1]{#1}

\newcommand{\nMuscles}{356\xspace}
\newcommand{\nDofs}{37\xspace}
\newcommand{\nMyoMuscles}{354\xspace}
\newcommand{\nWorlds}{4{,}096\xspace}
\newcommand{\nFibIter}{six\xspace}
\newcommand{\nFibIterRef}{40\xspace}
\newcommand{\rFwdMus}{\pending{0.88}\xspace}%
\newcommand{\rGradMus}{\pending{0.38}\xspace}%
\newcommand{\xMjwMus}{\pending{37}\xspace}%
\newcommand{\tFwdGone}{1.21\xspace}
\newcommand{\tGradGone}{1.86\xspace}
\newcommand{\tFwdNabla}{1.29\xspace}
\newcommand{\tGradNabla}{1.95\xspace}
\newcommand{\tRunFour}{\pending{12}\xspace}
\newcommand{\tMsmFour}{17.8\xspace}%
\newcommand{\rtfBatchForward}{\pending{8{,}800}\xspace}
\newcommand{\rtfBatchGradient}{\pending{3{,}800}\xspace}
\newcommand{\kitReturnTarget}{0.83\xspace}
\newcommand{\pMmCross}{\pending{93.5}\xspace}
\newcommand{\tMmCross}{\pending{14.2}\xspace}
\newcommand{\xMmTime}{\pending{1{,}300}\xspace}

\newcommand{\tnote}[1]{\textsuperscript{#1}}
\xspaceaddexceptions{\tnote}

\graphicspath{{figures/}}

\title{DiffNeuSim: A Differentiable Neuromechanical Simulator for Full-Body Musculoskeletal Motion}
\author{Hyoungseo Son and Seungmoon Song\thanks{Correspondence: s.song@northeastern.edu}\\
  {\normalsize Department of Mechanical and Industrial Engineering, Northeastern University, Boston, MA, USA}}
\date{}

\begin{document}
\maketitle

\begin{abstract}
Neuromechanical simulation can advance research in human biomechanics and motor control, but detailed models remain computationally expensive.
We present DiffNeuSim, a GPU-accelerated differentiable simulator of full-body musculoskeletal dynamics with \nMuscles Hill-type muscles, elastic tendons, skeletal dynamics and ground contact.
Across \nWorlds parallel worlds, one RTX 4090 runs simulations at approximately \rtfBatchForward times real time in aggregate, or \rtfBatchGradient times with gradients.
We used a common gradient-based policy learning algorithm and neural network architecture across all muscle-driven tasks.
Imitation policies learned diverse recorded motions from scratch within minutes, and generalist imitation achieved \pMmCross\% success in \tMmCross\,min, surpassing a model-free reinforcement learning benchmark with an estimated \xMmTime-fold training speedup while matching its tracking accuracy.
Walking and running emerged within about 18--30\,min without prescribed gait patterns or engineered control structures, enabling an LLM-assisted search of 42 reward settings within half a day.
Predictive locomotion reproduced human-like walking kinematics, kinetics and muscle activations, a U-shaped walking cost-of-transport curve, and near-human optimal walking and energetic crossover speeds.
Comparisons with rigid-tendon models highlighted the importance of modelling tendon elasticity for predicting walking gait, sprinting speed and jumping height.
DiffNeuSim provides an open-source framework for investigating the biomechanical and neural mechanisms of human movement.
\end{abstract}

\section*{Introduction}

Neuromechanical simulation couples models of motor control with physics-based models of the musculoskeletal system and environment to analyse movement and predict how it changes with task goals, impairments, and interventions.
By tracking or reproducing experimentally observed movements, simulations enable the estimation of otherwise difficult-to-measure quantities, including muscle excitations, muscle forces, and joint contact loads, providing insights into the neuromuscular mechanisms of movement~\cite{thelen2006computed,pizzolato2015ceinms}.
Predictive simulations further enable ``what-if'' investigations by generating movements under altered physiological properties, control mechanisms, or task objectives.
Such simulations allow researchers to test hypotheses about motor control by comparing alternative control architectures and examining how their mechanisms and parameters shape movement~\cite{park2026spinal,ton2026unified}.
They also have provided insights into the mechanisms underlying ageing-related and pathological gait~\cite{song2018ageing,ong2019predicting,johnson2022hemiparetic,bruel2022investigation}, enabled the development of assistive-device controllers~\cite{luo2024experiment,barati2026endtoend,robbins2026myoassist}, and supported the exploration of training strategies and personalized rehabilitation interventions~\cite{vanwouwe2024simulation,hammond2025neuromusculoskeletal}.

Computational speed influences what neuromechanical simulation studies are feasible and how quickly their findings can inform decisions.
Testing motor-control hypotheses often requires exploring alternative control architectures, objectives and parameter settings through repeated simulations and optimizations~\cite{park2026spinal,ton2026unified,pugliese2026connectome}.
For example, reproducing human-like walking from task-level objectives alone remains challenging~\cite{kim2023learning}, and investigating the conditions under which human-like gait emerges may require extensive exploration of alternative motor-control objectives.
Understanding movement impairments similarly requires systematically exploring how changes in physiological properties and control mechanisms affect movement~\cite{song2018ageing,ong2019predicting,johnson2022hemiparetic}.
For assistive-device development and rehabilitation, faster simulation enables rapid exploration of device designs, control strategies and interventions and could support real-time control~\cite{sartori2026ceinms} and personalized treatment planning~\cite{cashaback2024nsf}.

Recent advances have dramatically accelerated simulation and policy learning in robotics, opening new opportunities for biomechanically detailed musculoskeletal simulation.
GPU-accelerated physics engines enable thousands of simulated worlds to run in parallel~\cite{makoviychuk2021isaac,mjwarp}, while gradients of simulated dynamics can substantially accelerate policy learning~\cite{xu2022shac,xing2025sapo}.
For example, NablaEngine enables humanoid motion-tracking policy learning through analytically derived gradients of rigid-body dynamics~\cite{lu2026nablamimic}, whereas SuperMimic accelerates policy learning through a learned differentiable world model~\cite{liao2026supermimic}.
These advances have recently extended to musculoskeletal simulation, but existing full-body systems such as MyoSuite, MuscleMimic, and muscle-supermimic use simplified muscle-tendon dynamics by assuming rigid tendons~\cite{caggiano2022myosuite,li2026towards,he2026musclesupermimic}.
Tendon elasticity influences muscle force generation and the storage and return of mechanical energy, and neglecting it may affect the movements predicted by musculoskeletal simulations~\cite{zajac1989muscle,millard2013flexing}.
Among simulators that model elastic tendons, Hyfydy is neither GPU-accelerated nor differentiable~\cite{hyfydy}, while Bolt supports GPU-parallel simulation but does not provide simulator gradients for policy learning in its released implementation~\cite{wang2026bolt}.
Further discussion and comparison of related systems and learning approaches are provided in Supplementary Note~1.

Here we present DiffNeuSim, a GPU-accelerated differentiable neuromechanical simulator for full-body musculoskeletal models.
We also demonstrate a common gradient-based learning framework for individual-motion imitation, generalist imitation and task-driven predictive simulation.
We further investigate the importance of modelling tendon elasticity in predicting human movement.
Our contributions are:
\begin{itemize}
  \item A GPU-accelerated neuromechanical simulator with analytic gradients for full-body musculoskeletal models with \nMuscles Hill-type muscles and elastic tendons, achieving approximately \xMjwMus-fold higher forward throughput and \pending{16}-fold higher throughput with gradients than MuJoCo Warp's forward simulation across \nWorlds parallel environments on an RTX 4090;
  \item Integration of a common gradient-based learning framework for individual-motion imitation, generalist imitation with specialist fine-tuning, and task-driven predictive simulation. Generalist imitation exceeds MuscleMimic's reported success rate~\cite{li2026towards} with an estimated \xMmTime-fold speedup in training while matching its reported site error;
  \item Task-driven predictive walking and running without engineered control structures or prescribed gait patterns. The predicted walking gait achieves biomechanical agreement with human data comparable to biologically inspired control approaches~\cite{song2015neural,ogihara2001generation,aoi2019neuromusculoskeletal,di2023investigating,ton2026unified}, and the cost-of-transport curves of walking and running cross near the typical human
  walk-to-run transition speed;
  \item An investigation of the importance of modelling tendon elasticity, showing that rigid-tendon simplification alters predicted walking and reduces sprinting speed and jumping height;
  \item An open-source release of DiffNeuSim and the computational experiments presented in this study:
  \begin{itemize}
    \item Simulator and musculoskeletal models: \url{https://github.com/neumovelab/diffneusim}
    \item Learning frameworks, experimental code and results: \url{https://github.com/neumovelab/diffneusim-experiments}
  \end{itemize}
\end{itemize}

\section*{Results}

Additional methodological details and results are provided in Supplementary Notes~2--13.
The simulated movements are demonstrated in the accompanying video (\url{https://www.youtube.com/watch?v=aaMw1Ycz4FE}).

\subsection*{The simulator: speed, gradients and verification}

\subsubsection*{DiffNeuSim overview and performance}

\begin{figure*}[!b]%
  \centering
  \includegraphics[scale=0.92]{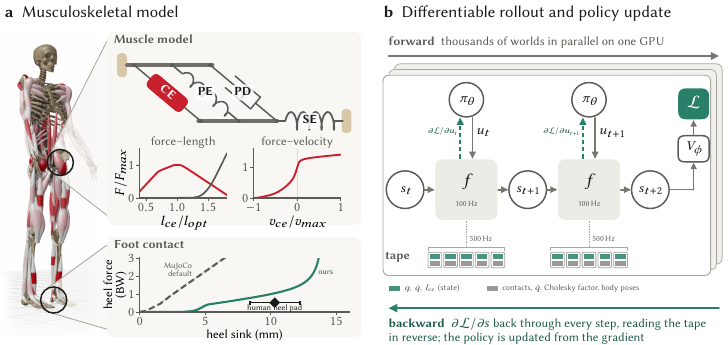}\hfill
  \includegraphics[scale=0.92]{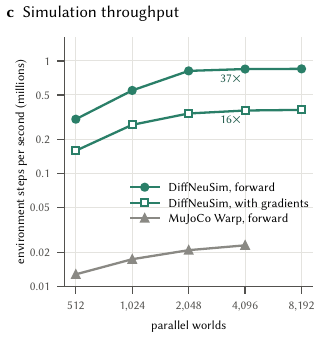}
  \caption{\textbf{DiffNeuSim overview.} \textbf{a,} Full-body musculoskeletal model with Hill-type muscles and elastic tendons, showing representative muscle force--length and force--velocity relationships and the foot--ground contact model compared with MuJoCo's default contact and measured human heel-pad mechanics.
  \textbf{b,} Differentiable simulation and policy learning across parallel environments on a GPU. The backward pass propagates gradients through the simulated dynamics and across time using information stored during the forward pass.
  \textbf{c,} Simulation throughput with and without gradient computation compared with MuJoCo Warp's forward simulation on an RTX 4090.}
  \label{fig:platform}
\end{figure*}

DiffNeuSim simulates and differentiates full-body musculoskeletal movement across thousands of parallel environments on a GPU (Fig.~\ref{fig:platform}).
At each time step, muscle excitations drive activation and Hill-type muscle--tendon dynamics, generating forces that act on the skeleton through rigid-body dynamics and ground contact.
The backward pass propagates derivatives through these processes and across time, providing gradients of a training objective or movement outcome with respect to earlier muscle excitations (Fig.~\ref{fig:platform}b).
The differentiable rigid-body engine builds on NablaEngine~\cite{lu2026nablamimic}, and the contact implementation follows MuJoCo~\cite{todorov2012mujoco}.

DiffNeuSim provides two musculoskeletal models: a full-body model with \nMuscles muscles, 37 bodies and \nDofs degrees of freedom (Fig.~\ref{fig:platform}a), and a reduced model with \pending{290} trunk and lower-limb muscles, 17 bodies and 23 degrees of freedom.
Both models are derived from the skeleton and muscle paths of MyoFullBody~\cite{li2026towards}, with muscle parameters fitted to corresponding OpenSim models~\cite{rajagopal2016fullbody}.
We further adapted the models to better represent human musculoskeletal mechanics.
In particular, we calibrated Achilles tendon compliance to better represent measured human tendon properties~\cite{kharazi2021achilles} and adjusted the foot--ground contact model to produce a compliant response at initial contact that progressively stiffens with indentation, consistent with the mechanical behaviour of the human heel pad~\cite{wearing2014force} (Fig.~\ref{fig:platform}a).
Both models support the Hill-type muscle formulations of Millard et al.~\cite{millard2013flexing} and Geyer and Herr~\cite{geyer2010muscle}, each with elastic or rigid
tendons.
Unless otherwise stated, the musculoskeletal results presented here use the Millard et al.\ formulation with elastic tendons.
Rigid tendons were used to investigate the effects of tendon elasticity on movement.

Across \nWorlds parallel worlds on one RTX 4090, DiffNeuSim simulated \rFwdMus million environment steps per second, equivalent to \rtfBatchForward times real time in aggregate, or \rGradMus million steps per second with gradients (Fig.~\ref{fig:platform}c).
Its forward throughput was \xMjwMus times that of MuJoCo Warp~\cite{mjwarp} simulating MuscleMimic's rigid-tendon MyoFullBody at the same batch size; even with gradients, DiffNeuSim achieved \pending{16} times MuJoCo Warp's forward throughput.
We verified the single-precision gradients against central finite differences of a separate double-precision implementation.
With and without ground contact, the median relative errors were on the order of $10^{-5}$ over one environment step and remained below $10^{-4}$ over sixteen steps.

\subsubsection*{Agreement with MuJoCo on the Unitree G1}
DiffNeuSim's rigid-body and contact dynamics closely matched those of MuJoCo on the Unitree G1 humanoid robot, with rounding-level agreement in single-step predictions and zero-shot
policy transfer with nearly identical tracking performance.
From the same state and input, one 20\,ms control step left the two simulators' joint positions within single-precision rounding of each other, with and without ground contact.
Replaying the same controls without feedback for 3\,s left their root positions \pending{0.31}\,mm apart, comparable to the drift caused by rounding-level perturbations within DiffNeuSim alone.
Two policies trained through DiffNeuSim's gradients to track a side kick without domain randomization completed all \pending{32} episodes from perturbed initial states in MuJoCo without modification, with nearly identical tracking errors in the two simulators.
On one RTX 4090, DiffNeuSim's G1 step took \tFwdGone\,ms forward and \tGradGone\,ms with gradients, compared with NablaEngine's \tFwdNabla\,ms and \tGradNabla\,ms, respectively.

\begin{figure*}[p]%
  \centering
  \makebox[\textwidth][l]{\scriptsize\textbf{a}\quad One-clip imitation from scratch}\par\smallskip
  \includegraphics[width=\textwidth]{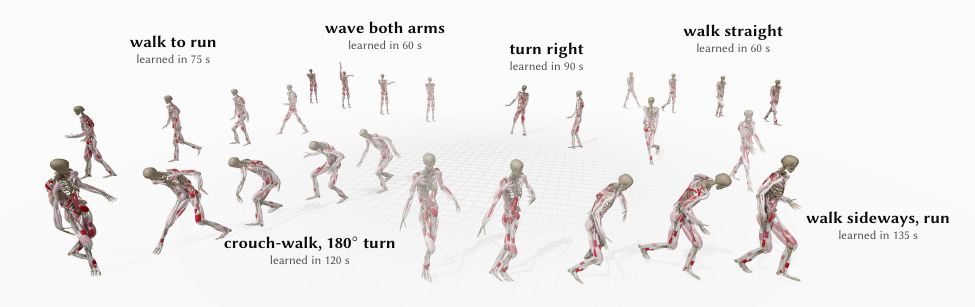}\par\medskip
  \begin{minipage}[t]{2.7in}\vspace{0pt}
    \makebox[2.7in][l]{\scriptsize\textbf{b}\quad Held-out return}\par\smallskip
    \includegraphics{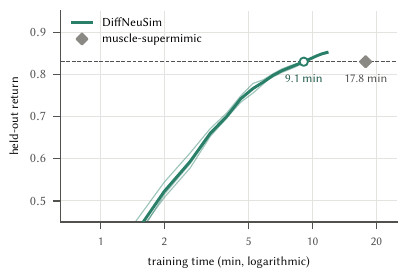}%
  \end{minipage}\hfill
  \begin{minipage}[t]{2.7in}\vspace{0pt}
    \makebox[2.7in][l]{\scriptsize\textbf{c}\quad Success on held-out clips}\par\smallskip
    \includegraphics{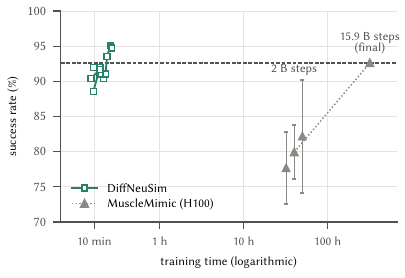}
  \end{minipage}\par\medskip
  \makebox[\textwidth][l]{\scriptsize\textbf{d}\quad Specialists fine-tuned from the generalist}\par\smallskip
  \includegraphics[width=\textwidth]{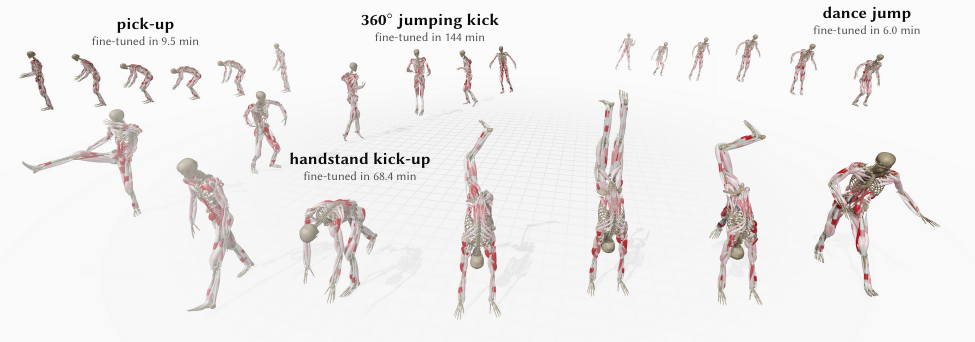}%
  \caption{\textbf{Individual-motion and generalist imitation learning.}
  \textbf{a,} Representative motions imitated by individual policies trained from scratch, with their training times.
  \textbf{b,} Held-out return of generalist imitation policies as a function of training time, compared with muscle-supermimic's learned-world-model approach on the same GPU (RTX 4090). Thin lines represent three training seeds and the thick line represents their mean.
  \textbf{c,} Success rates on held-out locomotion clips compared with MuscleMimic's reported results under its evaluation protocol. MuscleMimic's training times are estimated from its reported computational throughput.
  \textbf{d,} Challenging dynamic motions imitated by specialist policies fine-tuned from a generalist policy, with their fine-tuning times.
  In \textbf{a} and \textbf{d}, muscles are coloured by activation from light pink (0) to deep red (1).
  The motions in \textbf{a} and \textbf{d} are shown in the accompanying video.}
  \label{fig:imitation}
\end{figure*}

\subsection*{Learning muscle-driven movement}

Across all muscle-driven learning tasks, we used soft analytic policy optimization (SAPO)~\cite{xing2025sapo} with the same fully connected network architectures for the actor (1,024, 1,024 and 512 hidden units) and two critics (512 and 512 hidden units each).
The actor directly controls individual muscle excitations, and the policies are optimized through DiffNeuSim's gradients.

\subsubsection*{Individual-motion imitation learning}
Gradient-based policy learning in DiffNeuSim enabled policies to learn individual motions from scratch within minutes (Fig.~\ref{fig:imitation}a).
Policies successfully learned 14 of 20 tested motions by completing the reference motion without triggering termination (including a pelvis position-error threshold of 1\,m), with an average training time of 77\,s per successful motion on an RTX 4090.
The 14 policies achieved a mean site error (root mean square error of pelvis-relative positions of 16 body sites) of 5.3\,cm at their first successful checkpoints.

\subsubsection*{Generalist imitation learning and specialist fine-tuning}

Gradient-based policy learning in DiffNeuSim trained a generalist full-body motion-tracking policy in minutes, achieving accuracy comparable to conventional reinforcement learning that required hundreds of GPU-hours.
We trained three policies with different random seeds on MuscleMimic's imitation-learning benchmark~\cite{li2026towards}, comprising 972 training and 108 held-out motion clips from the KIT Whole-Body Human Motion Database~\cite{mandery2015kit}.
On one RTX 4090, the mean held-out return reached \kitReturnTarget after 9.1\,min, approximately twice as fast as muscle-supermimic~\cite{he2026musclesupermimic}, which learns through a differentiable world model and required \tMsmFour\,min to reach
the same return on the same GPU (Fig.~\ref{fig:imitation}b).
Under MuscleMimic's evaluation protocol, after \tRunFour\,min of training, the three policies achieved a mean success rate of 91.6\% and a mean site error of 2.28\,cm, matching MuscleMimic's reported site error.
With further training, one policy reached \pMmCross\% success after \tMmCross\,min, exceeding MuscleMimic's reported 92.62\% after 15.87 billion environment steps, corresponding to an estimated \pending{320} GPU-hours at its reported throughput (Fig.~\ref{fig:imitation}c).

Gradient-based fine-tuning in DiffNeuSim enabled specialist policies to reproduce challenging dynamic motions, including jumps, handstands and aerial rotations, at normal muscle strength (Fig.~\ref{fig:imitation}d).
Starting from the generalist policy trained on an RTX 4090, we fine-tuned separate policies on individual motion clips from the AMASS and KIT datasets~\cite{mahmood2019amass,mandery2015kit}, using an H200 whose larger GPU memory enabled longer backpropagation windows.
The specialists completed a dance into a jump after \pending{6.0}\,min, a pick-up after \pending{9.5}\,min and a handstand kick-up after \pending{68.4}\,min of fine-tuning.
A kick with a $360^\circ$ jumping twist required \pending{2.4}\,h of fine-tuning (\pending{7,620} epochs), after which the policy completed the entire motion.
The four specialist policies achieved a mean site error of 3.0\,cm across their respective motions.
For comparison, MuscleMimic reports approximately one billion environment steps of fine-tuning for challenging dynamic motions~\cite{li2026towards}, corresponding to an estimated 20 GPU-hours on one H100, which shares the H200's GPU architecture.

We further demonstrated zero-shot transfer of muscle-driven policies, trained in DiffNeuSim without domain randomization, to MuJoCo equipped with our elastic-tendon muscle implementation.
The generalist policy achieved nearly identical tracking performance across all 108 held-out motions in both simulators, and the specialist policies reproduced challenging dynamic motions after transfer (Supplementary Note~9).

\subsubsection*{Task-driven predictive policies for walking and running}

\begin{figure}[!t]%
  \centering
  \begin{minipage}[t]{3.4in}\vspace{0pt}
    {\scriptsize\textbf{a}\quad Cost of transport across locomotion speeds}\par
    \includegraphics{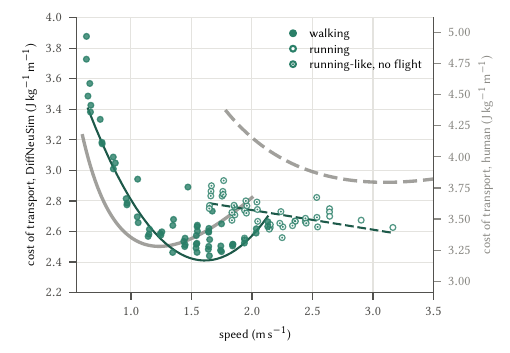}\par\medskip%
    {\scriptsize\textbf{b}\quad Predicted walking and running gaits}\par\smallskip
    \makebox[3.4in][c]{\includegraphics[width=3.06in]{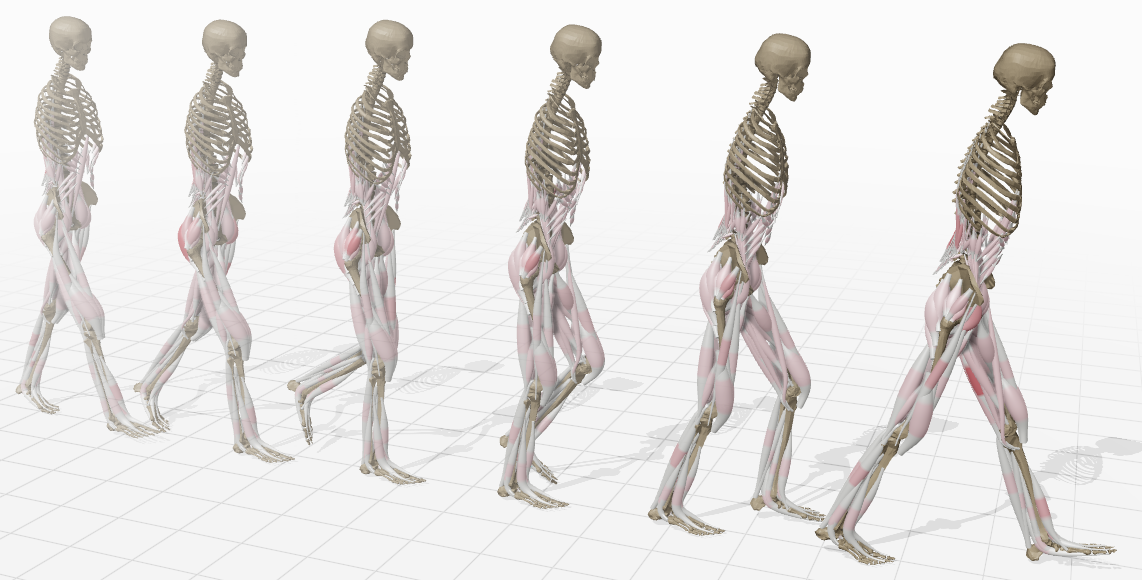}}\par
    \makebox[3.4in][c]{\fontsize{6}{7}\selectfont walking}\par\medskip
    \makebox[3.4in][c]{\includegraphics[width=3.06in]{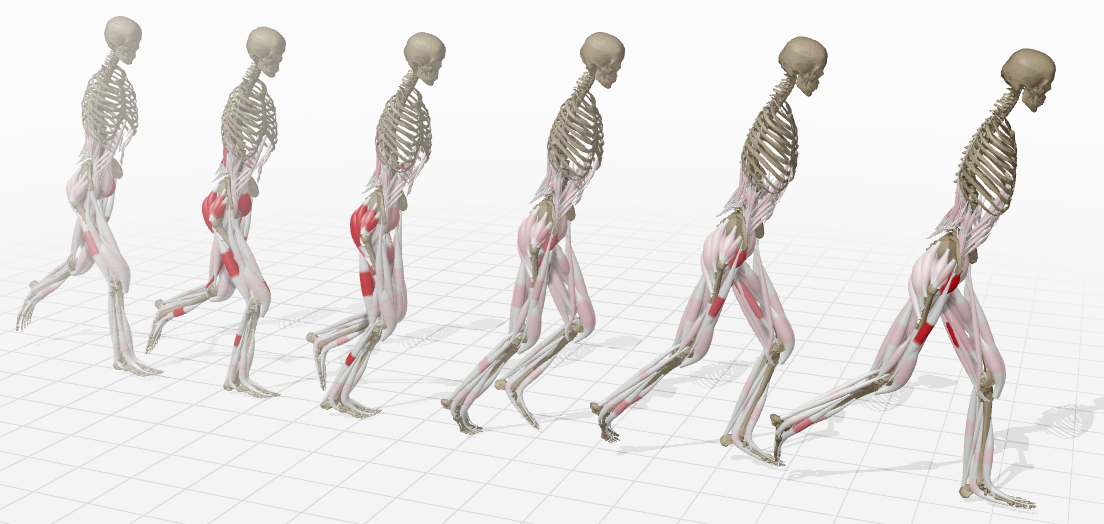}}\par
    \makebox[3.4in][c]{\fontsize{6}{7}\selectfont running}
  \end{minipage}
  \caption{\textbf{Task-driven predictive walking and running.}
  \textbf{a,} Metabolic cost of transport as a function of locomotion speed. Filled circles represent walking policies, open circles represent running policies with flight phases, and open circles with a central dot represent running-like policies without flight phases. Green curves represent a quadratic fit to the lowest walking costs and a linear fit to the costs of running and running-like gaits. Grey curves represent human walking and running data~\cite{ralston1958energy,batliner2018metabolic}, plotted against the right vertical axis. \textbf{b,} Stroboscopic visualization of representative walking and running gaits learned through DiffNeuSim's gradients. The gaits in \textbf{b} are shown in the accompanying video.}
  \label{fig:cot}
\end{figure}

Gradient-based policy learning in DiffNeuSim produced sustained walking and running across a range of speeds without engineered control structures or prescribed gait patterns (Fig.~\ref{fig:cot}).
Training on one RTX 4090 took approximately \pending{18}\,min for walking and \pending{30}\,min for running.
The reward encouraged tracking a commanded forward speed while penalizing muscle activation, metabolic cost, excessive head acceleration, centre-of-mass height below a threshold and falls.
We explored the reward weights through an LLM-assisted heuristic search of 42 candidates to produce joint kinematics resembling human walking at 1.3\,m/s.
Keeping the selected reward settings fixed, we then trained policies at commanded speeds from 0.7 to 4.0\,m/s with multiple random seeds.
Walking emerged over a speed range of \pending{0.63--2.13}\,m/s, and running and running-like gaits emerged over 1.65--3.16\,m/s, with the overlapping ranges indicating
that gait emergence depended on the random seed and training procedure.
Following common approaches~\cite{carrier1984energetic,falisse2019rapid,leem2026exo}, we fitted a quadratic curve to the minimum walking cost of transport (COT) at each commanded speed and a linear function to the COT of running and running-like gaits (Fig.~\ref{fig:cot}a).
The quadratic fit predicted an energetically optimal walking speed of \pending{1.61}\,m/s, somewhat faster than typical preferred human walking speeds, and the two fitted curves intersected at \pending{2.15}\,m/s, near the typical human walk-to-run transition
speed.

\begin{figure*}[!t]%
  \centering
  \includegraphics{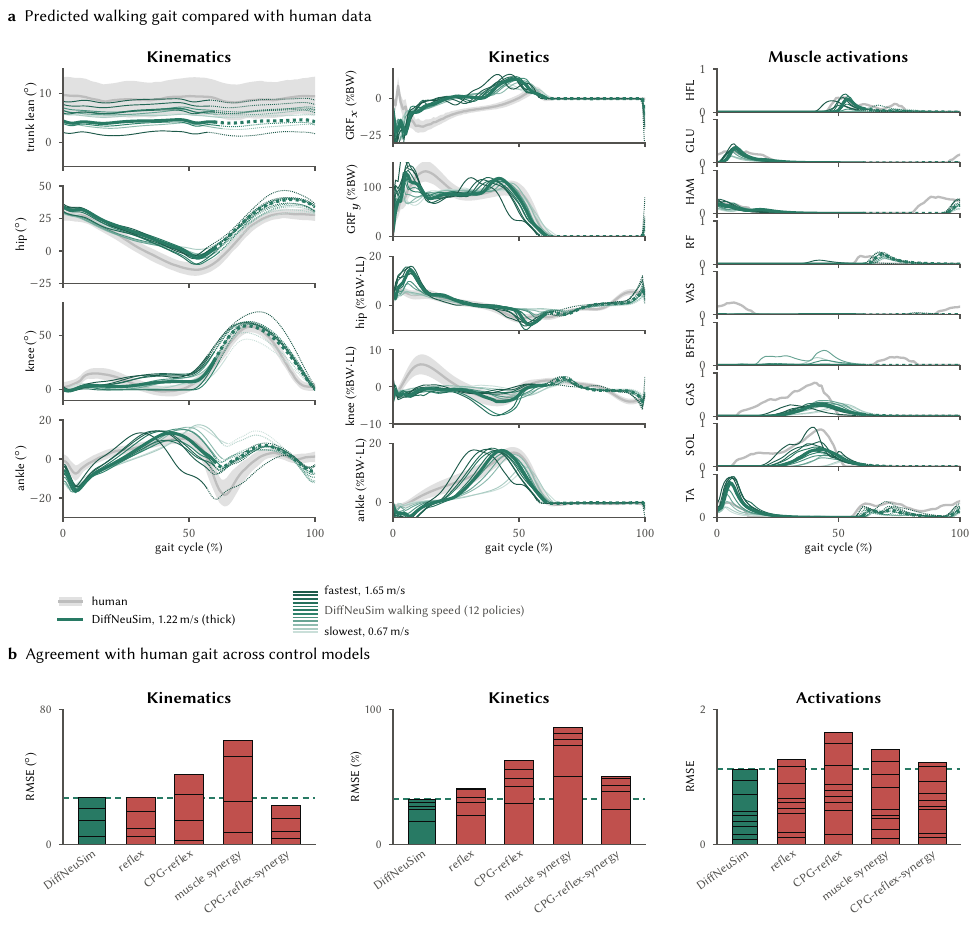}
  \caption{\textbf{Predicted walking gait.}
  \textbf{a,} Predicted walking gait of DiffNeuSim compared with human data, including joint kinematics, ground reaction forces, joint moments and muscle activation patterns over the gait cycle. Grey curves and bands indicate the human mean and standard deviation, and green curves indicate DiffNeuSim policies, colour-coded by walking speed. The thick green curve represents the policy walking at 1.22\,m/s.
  \textbf{b,} Summed root mean square errors (RMSEs) of kinematics, kinetics and muscle activations relative to human data for DiffNeuSim and four biologically inspired control models reported previously~\cite{ton2026unified}: reflex-based~\cite{song2015neural}, CPG-reflex-based~\cite{ogihara2001generation}, muscle-synergy-based~\cite{aoi2019neuromusculoskeletal}, and CPG-reflex-synergy-based~\cite{di2023investigating}. Green bars show DiffNeuSim and red bars show the four biologically inspired control models. Lower values indicate closer agreement with human gait. Each bar stacks the RMSEs of individual variables in the order
shown in \textbf{a}, from top to bottom.}
  \label{fig:predictive}
\end{figure*}

The predictive walking policy reproduced key features of human walking at a typical walking speed, with overall agreement comparable to that of established biologically inspired control models~\cite{song2015neural,ogihara2001generation,aoi2019neuromusculoskeletal,di2023investigating}.
At a commanded speed of 1.3\,m/s, the policy walked at an average speed of \pending{1.22}\,m/s, with \pending{117} steps per minute and a stride length of \pending{1.25}\,m.
Its joint kinematics, ground reaction forces, joint moments and muscle activations broadly agreed with human walking data (Fig.~\ref{fig:predictive}a).
The aggregate errors were comparable to those of the best-performing biologically inspired control models, with lower errors in kinetics and muscle activations than those of all four previously reported controllers~\cite{ton2026unified} (Fig.~\ref{fig:predictive}b).
However, the policy exhibited insufficient knee flexion during early stance and excessive flexion during mid-stance, accompanied by almost no activation of the vasti, the primary knee extensors.

Running and running-like gaits emerged with and without flight phases, respectively.
Between \pending{1.65} and approximately \pending{2.5}\,m/s, policies exhibited running-like gaits without flight phases, with in-phase oscillations of the potential and forward kinetic energies of the centre of mass~\cite{falisse2019rapid}.
At speeds below approximately \pending{2.0}\,m/s, however, the centre of mass still rose over the stance leg, as in walking~\cite{cavagna1977mechanical}.
At higher speeds, policies exhibited running with flight phases, reaching \pending{4.86}\,m/s with a minor modification to the reward.
Detailed biomechanics of predictive running are presented in Supplementary Fig.~S6.

\subsection*{Effects of tendon elasticity on predicted movement}

\begin{figure*}[!b]%
  \centering
  \begin{minipage}[t]{4.55in}\vspace{0pt}
    {\scriptsize\textbf{a}\quad Effects of tendon elasticity on walking gait\strut}\par\nointerlineskip\vspace{0.04in}
    \includegraphics[trim=0 0 0 0.163in, clip]{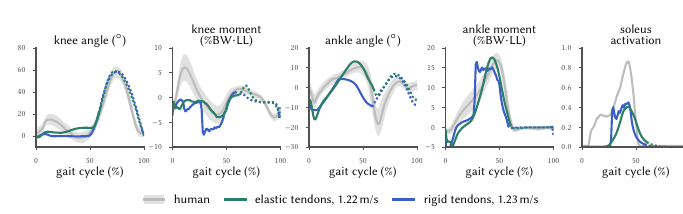}\par\nointerlineskip\vspace{0.08in}
    {\scriptsize\textbf{b}\quad Sprinting with elastic and rigid tendons\strut}\par\nointerlineskip\vspace{\dimexpr0.04in-0.002in\relax}
    \includegraphics{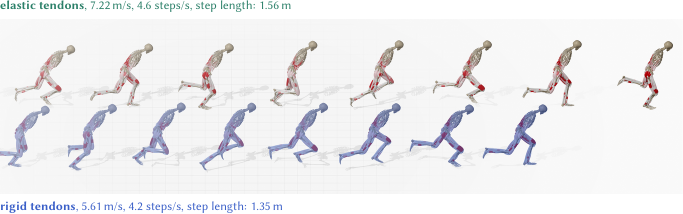}
  \end{minipage}\hfill
  \begin{minipage}[t]{1.692in}\vspace{0pt}
    {\scriptsize\begin{tabular}[t]{@{}l@{\quad}l@{}}\textbf{c} & Vertical jumping with\\ & elastic and rigid tendons\end{tabular}}\par\nointerlineskip\vspace{0.04in}
    \includegraphics{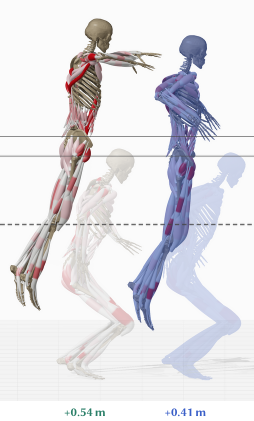}
  \end{minipage}
  \caption{\textbf{Effects of tendon elasticity on predicted movement.}
  \textbf{a,} Selected kinematics, kinetics and muscle activations during walking with elastic and rigid tendons at the same commanded speed of 1.3\,m/s, compared with human gait data (grey).
  \textbf{b, c,} Body configurations during sprinting (\textbf{b}) and maximum-height jumping (\textbf{c}) with elastic and rigid tendons. In \textbf{b}, configurations are shown at equal time intervals and actual spatial positions; in \textbf{c}, the crouched and apex configurations and the rise of the centre of mass above standing height are shown. Models with rigid tendons are tinted blue.
  The motions in \textbf{b} and \textbf{c} are shown in the accompanying video.}
  \label{fig:tendon}
\end{figure*}

To investigate the importance of modelling tendon elasticity in predicting movement, we trained walking, sprinting and jumping policies with elastic and rigid tendons using the same task-specific reward settings.
Tendon elasticity substantially affected the predicted walking gait.
At the commanded walking speed of 1.3\,m/s, both models walked at nearly the same average speed (1.23 and 1.22\,m/s with rigid and elastic tendons, respectively), but the rigid-tendon model took fewer, longer steps than the elastic-tendon model and human walking (cadence: 103, 117 and 123 steps/min; stride length: 1.44, 1.25 and 1.28\,m, respectively, for rigid tendons, elastic tendons and human data~\cite{fukuchi2018public}).
The rigid-tendon model also exhibited altered joint kinematics and greater deviations in kinetics from human walking data (Fig.~\ref{fig:tendon}a).
In particular, its knee remained at the extension limit for approximately 45\% of the gait cycle, and its peak ankle plantarflexor moment during push-off was lower.

The rigid-tendon model also showed reduced performance in dynamic movements, achieving lower sprinting speeds (5.61 versus 7.22\,m/s; Fig.~\ref{fig:tendon}b) and jumping heights (0.413 versus 0.536\,m, best of three training runs per tendon variant; Fig.~\ref{fig:tendon}c).

\section*{Discussion}

DiffNeuSim demonstrates that efficient gradient-based learning can be achieved in biomechanically detailed musculoskeletal models with elastic tendons.
By propagating gradients through muscle--tendon dynamics, skeletal dynamics and ground contact, it enables controllers to learn directly from the simulator's derivatives rather than relying solely on model-free reinforcement learning or learned dynamics models.
We demonstrated rapid learning of individual-motion and generalist imitation policies, the emergence of human-like locomotion from task-level objectives, and the importance of modelling tendon elasticity in predicting movement.
Together, these findings highlight the potential of differentiable neuromechanical simulation for investigating the neural and biomechanical mechanisms of human movement.

Training muscle-driven policies in minutes makes it practical to explore a substantially wider range of motor-control hypotheses, physiological conditions and simulation settings.
Preliminary experiments suggest that further gains in learning efficiency are possible: shortening the backpropagation window from 64 to 32 steps allowed us to double the number of parallel environments, reducing predictive walking training time to about 10\,min on an RTX 4090.
Rapid policy learning makes automated experimentation practical by allowing LLM agents to iteratively propose and evaluate changes to learning objectives and training configurations~\cite{choi2026instantmimic}.
For instance, our LLM-assisted reward-tuning experiment evaluated 42 reward settings for predictive walking in approximately 11 hours, with each policy trained from scratch and evaluated against human gait data.
Such exploration would be considerably more computationally demanding with conventional model-free reinforcement learning of full-body muscle-driven controllers~\cite{li2026towards}.
Together, rapid policy learning and automated experimentation could enable more systematic investigations of how motor-control objectives and physiological properties shape human movement.

One particularly valuable application of rapid reward exploration is to investigate how locomotion emerges under minimal assumptions about gait patterns and neuromuscular control.
Previous learning-based approaches have facilitated muscle-driven locomotion by prescribing gait-specific features, such as step lengths or reference poses~\cite{park2022generative,leem2026exo,gambietz2026closing}, or by incorporating specialized control and exploration mechanisms~\cite{schumacher2025emergence,park2026learning,lee2026lambda}.
Although direct muscle control using functional objectives has enabled locomotion learning, including impressive sprinting performance in recent work~\cite{wang2026bolt}, reproducing human-like walking remains challenging~\cite{kim2023learning}.
Using a general-purpose learning algorithm and a multilayer perceptron architecture shared across imitation and predictive tasks, policies directly controlling individual muscles learned sustained walking and running with task-level objectives and general effort and stability penalties.
Notably, the predicted walking gait reproduced characteristic features of human joint kinematics, kinetics and muscle activation patterns with overall agreement comparable to that of the best-performing biologically inspired control models, which use sagittal-plane musculoskeletal models
with 18 muscles~\cite{ton2026unified}.
Furthermore, the predicted cost of transport reproduced the characteristic U-shaped relationship with walking speed and crossed that of running and running-like gaits near the typical human walk-to-run transition speed.
To our knowledge, this is among the first demonstrations of this level of biomechanical agreement in predictive walking with direct muscle control and without engineered neuromuscular control structures or gait patterns prescribed in the training reward.

Our findings also highlight the importance of biomechanical fidelity in both imitation and predictive neuromechanical simulations.
Replacing elastic with rigid tendons, a common simplification in computationally efficient musculoskeletal simulators, altered the predicted walking gait and reduced sprinting speed and jumping height.
Other biomechanical properties, including foot--ground contact mechanics, muscle paths, passive joint properties and wobbling mass dynamics, may also substantially influence simulated movement.
Although imitation policies trained with simplified musculoskeletal models may reproduce observed motion sufficiently well for applications such as character animation, inaccuracies in the underlying mechanics may bias estimates of muscle activations, muscle forces and joint loads.
In predictive simulations, reward formulations and training procedures tuned for simplified models may compensate for inaccuracies in their mechanics, potentially producing plausible movements for the wrong reasons.
Improvements in control optimization should therefore be accompanied by careful evaluation of the underlying musculoskeletal model.

DiffNeuSim currently has limitations in biomechanical fidelity and model flexibility, and extending its capabilities requires careful algorithmic choices to maintain differentiability and GPU efficiency.
For example, muscle paths are currently represented by via points rather than wrapping surfaces, which can introduce errors in muscle moment arms, particularly at large joint excursions.
Furthermore, simulation algorithms involving complex collision geometries and variable contact configurations, as encountered in locomotion over irregular terrain and dexterous object manipulation, may be less amenable to efficient GPU parallelization.
Extending DiffNeuSim's modelling capabilities while maintaining efficient simulation and differentiation remains an important direction for future development.

We envision DiffNeuSim as complementary to, rather than a replacement for, established musculoskeletal simulation ecosystems such as OpenSim~\cite{seth2018opensim,dembia2020opensim,pizzolato2015ceinms,hammond2025neuromusculoskeletal,geijtenbeek2019scone} and MyoSuite~\cite{caggiano2022myosuite,wang2026myochallenge,robbins2026myoassist,li2026towards}.
One way these platforms can complement each other is by separating policy training from deployment.
Our zero-shot policy transfer to MuJoCo demonstrates the potential to leverage efficient gradient-based learning in DiffNeuSim while retaining established simulation environments for subsequent research and applications.
We see value in developing specialized simulators tailored to the scientific questions and computational requirements of individual research projects.
AI-assisted software development may make it increasingly practical to build and extend such research-specific tools.
Capabilities developed through these projects could subsequently be incorporated into established simulation ecosystems, benefiting the broader research community.
To support this complementary development and encourage reproducible research across platforms, we make DiffNeuSim's simulation and training code, musculoskeletal models, trained policies, and experimental results openly available (see Code availability).
\section*{Methods}

\subsection*{Neuromechanical simulation and numerical verification}
Further implementation details, musculoskeletal model parameters, calibration procedures and numerical verification are provided in Supplementary Notes~2--6.
Additional investigations of the muscle formulation and foot--ground contact parameters are presented in Supplementary Notes~12 and~13, respectively.

\subsubsection*{GPU-accelerated differentiable simulation architecture}
DiffNeuSim builds on the GPU execution model of NablaEngine~\cite{lu2026nablamimic}, assigning one warp of 32 threads to each simulated world. At each physics step, a muscle kernel computes muscle forces and their contributions to generalized forces, followed by two rigid-body kernels that compute unconstrained dynamics, collision and constraint data, solve the constraint problem, and integrate the state. The state comprises generalized positions and velocities, muscle activations and fibre lengths. The forward pass records intermediate quantities required by the analytic adjoint, which runs two kernels per physics step in reverse chronological order. The tape stores the input generalized positions, velocities and fibre lengths, active constraint rows, final accelerations, the Cholesky factor and body poses. Muscle and contact forces are recomputed during the backward pass. Warp communication follows a fixed order without floating-point atomic operations.

DiffNeuSim extends NablaEngine's architecture to muscle-driven multibody dynamics, integrating and differentiating muscle geometry, activation and elastic-tendon dynamics together with skeletal dynamics and contact.
Parts of our implementation are derived from NablaEngine's source code, including the tape and contact-row layouts, dense Cholesky factorization, exact line search, per-contact Hessian and forward kinematics.
We implemented the contact law, collision detection, constraint construction, solver warm start and stopping criteria based on the MuJoCo 3.16 C source code~\cite{todorov2012mujoco,mujocodocs}.
For implicit solves, the analytic adjoint differentiates the solution of the governing equations rather than the individual solver iterations.
The resulting derivatives are valid for converged solutions within fixed differentiable regimes, may differ from those of truncated solves and may be undefined at transitions between contact, constraint or muscle regimes.
The simulation kernels support both single and double precision, with single precision used for policy training.

\subsubsection*{Extensions to skeletal dynamics}
We implement the linear couplings of the spine and shoulder girdle as exact kinematic couplings, with the lumbar levels sharing a master coordinate as in the source model. The tibia and patella are fixed at their positions in a slightly flexed knee, rather than translated through MyoFullBody's quartic knee couplings; 18 quadriceps via points move as functions of knee angle. The hips are ball joints integrated using quaternions.
We implemented passive joint damping and linear toe joint springs with a stiffness of 25\,N\,m/rad following Falisse et al.~\cite{falisse2022}.
For the hips, we implemented exponential passive torques~\cite{riener1999} to resist motion beyond MyoFullBody's range-of-motion bounds along individual rotational axes, rather than using MuJoCo's single cone-angle limit for ball joints.
We also implemented ellipsoid--floor collision detection using MuJoCo's convex support function~\cite{mujocodocs}.
All these extensions have corresponding analytic adjoints.
The equations of motion are integrated using MuJoCo's Euler integrator with explicit damping. For the musculoskeletal experiments, each 10\,ms environment step comprises five 2\,ms physics steps, with muscle excitations held constant over the five steps.

\subsubsection*{Foot--ground contact mechanics}
Foot--ground contact and hinge limits follow MuJoCo's compliant constraint formulation~\cite{todorov2012mujoco,todorov2014convex,mujocodocs}, with impedance and reference acceleration determined by the \texttt{solimp} and \texttt{solref} parameters. Friction is represented by pyramidal friction cones, and the constraint problem is solved by at most six primal Newton iterations (tolerance $10^{-12}$, eight line-search iterations).
Rather than introducing a different contact law, we adjusted the foot--ground contact parameters to produce a compliant response at initial contact followed by progressive stiffening with indentation, consistent with measured human heel-pad mechanics~\cite{wearing2014force}.
The full-body configuration described in the engine documentation includes floor contact for 87 collision geometries (68 capsules and 19 ellipsoids) and self-contact for 68 capsule pairs. The contact adjoint implicitly differentiates the converged constraint solution with the contact set and active constraint regimes fixed.

To characterize the normal force response, we evaluated MuJoCo's closed-form static contact force as a function of vertical heel indentation in five heel-strike postures, using the selected floor parameters and MuJoCo's defaults (Fig.~\ref{fig:platform}a). The force was normalized by body weight. The selected parameters were chosen to give approximately 10\,mm of heel indentation at one body weight.

\subsubsection*{Elastic-tendon muscle dynamics}
We used the damped equilibrium Hill-type muscle model of Millard et al.~\cite{millard2013flexing}.
Let $l_{mtu}$, $l_{ce}$, $l_{opt}$ and $l_{slack}$ denote muscle--tendon length, fibre length, optimal fibre length and tendon slack length, respectively.
Normalized fibre length and velocity are $\tilde l=l_{ce}/l_{opt}$ and $\tilde v=(l_{ce}-l_{ce}^{-})/(h v_{max}l_{opt})$, where $l_{ce}^{-}$ is the previous fibre length, $h$ is the physics time step and $v_{max}$ is the maximum shortening velocity in optimal fibre lengths per second.
Pennation angle $\alpha$ follows the constant-thickness relation $l_{ce}\sin\alpha=l_{opt}\sin\alpha_{opt}$.
The normalized force equilibrium is
\begin{equation}
 f_T\!\left(\frac{l_{mtu}-l_{ce}\cos\alpha}{l_{slack}}\right)
 =\left[a f_L(\tilde l)f_V(\tilde v)+f_P(\tilde l)
 +\beta\tilde v\right]\cos\alpha,
 \label{eq:fibre}
\end{equation}
where $a$ is muscle activation, $\beta$ is the normalized fibre damping coefficient (0.1), and $f_L$, $f_V$, $f_P$ and $f_T$ are the active force--length, force--velocity, passive fibre force--length and tendon force--length curves, respectively.
Tendon force is $F=F_{max}f_T$, where $F_{max}$ is the maximum isometric fibre force.
We evaluate Millard's quintic B\'ezier curves from tables and impose the formulation's lower bound on fibre length.
Activation follows MuJoCo's dynamics~\cite{mujocodocs}, with time constants of 10 and 40\,ms for activation and deactivation, blended over a width of 0.1 in the difference between excitation and activation.

At each physics step, GPU threads compute muscle path lengths, moment arms, activations and tendon forces.
The moment arm of muscle $m$ about coordinate $q_j$ is $r_{mj}=-\partial l_m/\partial q_j$, computed in a single sweep along the muscle's kinematic chain.
We solve Eq.~\eqref{eq:fibre} for $l_{ce}$ using Newton's method, initialized from the previous fibre length, with \nFibIter iterations during training and \nFibIterRef in the double-precision reference.
Muscle forces contribute $\mathbf r_m F_m$ to the generalized forces, accumulated through fixed-order warp shuffles.
For the equilibrium residual $R(l_{ce};a,l_{mtu},l_{ce}^{-})$, the implicit function theorem gives
\begin{equation}
 \frac{\partial l_{ce}}{\partial\theta}
 =-\frac{\partial R/\partial\theta}
 {\partial R/\partial l_{ce}},
 \qquad \theta\in\{a,l_{mtu},l_{ce}^{-}\},
 \label{eq:fibreadjoint}
\end{equation}
evaluated at a converged equilibrium with $\partial R/\partial l_{ce}\ne0$.
The muscle adjoint propagates derivatives through fibre equilibrium, activation and muscle geometry.
DiffNeuSim also supports the Hill-type formulation of Geyer and Herr~\cite{geyer2010muscle}.

\subsubsection*{Musculoskeletal models and parameter calibration}
We constructed a full-body musculoskeletal model based on MyoFullBody, as distributed with MuscleMimic~\cite{li2026towards}, removing the finger joints and their muscles and merging bodies welded to their parents.
The resulting model is 1.69\,m tall and has a body mass of 84.3\,kg, with \nMuscles muscles, 37 bodies and \nDofs degrees of freedom, including six degrees of freedom for the
free root.
We added the bilateral abductor pollicis longus muscles to MuscleMimic's \nMyoMuscles muscles.
Muscle paths consist of straight segments connecting anatomical via points, with fitted via points replacing MyoFullBody's wrapping surfaces for 13 leg muscles per side.
We also constructed a reduced model for predictive locomotion with \pending{290} trunk and lower-limb muscles, 17 bodies and 23 degrees of freedom, with the mass of the arms incorporated into the trunk.
The two models share the same body height and total mass, as well as the anatomical and physiological parameters of corresponding muscles.

We obtained muscle parameters from established OpenSim models~\cite{seth2018opensim}: Rajagopal et al.~\cite{rajagopal2016fullbody}, with updates by Lai et al.~\cite{lai2017antagonist} and Uhlrich et al.~\cite{uhlrich2022retraining}, for the legs; Christophy et al.~\cite{christophy2012lumbar} for the trunk; MoBL-ARMS~\cite{saul2015mobl} for the upper limbs; and McFarland et al.~\cite{mcfarland2023hand} for seven wrist muscles per arm.
We retained the source models' muscle parameters, including pennation angles, and Millard's characteristic curves where available.
For lumbar muscles originally modelled using Thelen's formulation~\cite{thelen2003adjustment}, we fitted the characteristic curves of the Millard formulation to those of the source model.
Following previous neuromechanical simulation studies~\cite{jackson2017muscle,ton2026unified}, we increased the compliance of the tendons of the soleus and medial and lateral gastrocnemius to better represent experimentally measured Achilles tendon stretch during human locomotion~\cite{kharazi2021achilles}.

\subsubsection*{Computational benchmarks and numerical verification}
We measured simulation throughput as the number of parallel worlds divided by the elapsed time per environment step, with and without gradient computation.
Forward steps were replayed as CUDA graphs, and timings were reported as medians across three processes on an RTX 4090.
We benchmarked MuJoCo Warp~\cite{mjwarp} simulating MuscleMimic's rigid-tendon MyoFullBody at matched batch sizes and time steps.
DiffNeuSim used recorded states from held-out motions, whereas MuJoCo Warp started from its default pose with zero actions.
The comparison therefore includes differences in musculoskeletal models and simulation configurations.

We verified the single-precision analytic gradients against central finite differences of a separate double-precision CPU implementation.
We compared directional derivatives of the final state with respect to the initial state and muscle excitations over 1, 4 and 16 environment steps, with and without ground contact.
To evaluate gradients within fixed differentiable regimes, we retained perturbations that preserved the contact decisions and muscle-curve pieces of the unperturbed trajectories, independently of the derivative error.

We verified the rigid-body and contact dynamics against MuJoCo using the Unitree G1 humanoid robot with matched model and simulation settings.
We compared single-step state predictions with and without ground contact and the divergence of trajectories over 3\,s of open-loop replay.
We further evaluated zero-shot policy transfer by running side-kick policies trained in DiffNeuSim without domain randomization in MuJoCo from 32 perturbed initial states.

\subsection*{Learning muscle-driven movement}
Detailed training configurations, evaluation protocols and additional results for imitation learning, predictive learning and investigations of tendon elasticity are provided in Supplementary Notes~7--11.

\subsubsection*{Gradient-based policy optimization}
We trained neural-network policies with soft analytic policy optimization (SAPO)~\cite{xing2025sapo}, a maximum-entropy extension of short-horizon actor--critic learning~\cite{xu2022shac}. The actor controls each muscle independently through a state-dependent squashed Gaussian, with excitation $u_m=(1+\tanh(\mu_m+\sigma_m\epsilon_m))/2$, where $\epsilon_m$ is standard normal noise.
Across all muscle-driven learning tasks, we used the same fully connected network architectures, with three hidden layers of 1,024, 1,024 and 512 units for the actor and two hidden layers of 512 units each for both critics.
In each epoch, parallel environments are advanced for $H$ environment steps.
The actor maximizes the discounted entropy-augmented rewards over this window and a terminal value estimated by the mean of the twin critics, with gradients propagated through the simulated trajectories.
The critics are fitted to temporal-difference targets constructed using the minimum of their value estimates and trace parameter $\lambda$.
The entropy temperature is adjusted towards a target entropy.

Relative to the published SAPO recipe, the standard imitation configuration uses four rather than 16 critic passes per epoch, an initial entropy temperature of 0.1 rather than 1, and clipping of each environment's action gradient to twice the median norm across environments before global gradient clipping. We evaluate a Polyak average of the actor weights following NablaMimic~\cite{lu2026nablamimiccode}. Training wall-clock time includes the training loop and checkpoint saves but excludes evaluation, process initialization and the one-time engine build; compilation within the training loop is included, with a warm compilation cache. Training hyperparameters, including the number of parallel environments, backpropagation window, optimizer and learning-rate schedule, varied across tasks.

\subsubsection*{Individual-motion imitation}
We trained separate policies from scratch on 20 motion clips from the KIT Whole-Body Human Motion Database~\cite{mandery2015kit} and AMASS~\cite{mahmood2019amass}, converting MuscleMimic's retargeted motions to our full-body model.
We used MuscleMimic's imitation objective~\cite{li2026towards}, which includes tracking the positions of 17 body sites.
Episodes terminate when the mean site or root-position deviation exceeds 1\,m, the root-orientation deviation exceeds 1\,rad, the pelvis height falls below 0.6\,m or the trunk tilts more than $60^\circ$ from vertical.
We evaluated policies deterministically from the first reference frame and defined training time to success as the time of the first of three successive checkpoints completing the clip.
Tracking precision was quantified as the root mean square error of the pelvis-relative positions of 16 body sites, averaged over the evaluated time steps.
Training times and tracking errors were averaged over successful policies.

\subsubsection*{Generalist imitation and specialist fine-tuning}
We trained three generalist policies on MuscleMimic's benchmark of 972 training and 108 held-out KIT motion clips~\cite{li2026towards,mandery2015kit}, with 4,096 parallel environments and a backpropagation window of 32 steps on an RTX 4090.
Training episodes start from reference states sampled across the training clips.
The held-out return is the mean normalized cumulative reward across held-out clips, with zero reward assigned to frames after early termination~\cite{he2026musclesupermimic,liao2026supermimic}.
We estimated the training time to a return of \kitReturnTarget by interpolation between checkpoint evaluations and compared it with three runs of muscle-supermimic's BlinkMimic recipe on the same GPU at 100\,Hz.
For comparison with MuscleMimic's published results, we also evaluated policies using sampled actions with three sampling seeds per policy and 0.5\,m site and root-position termination bounds.
Tracking errors were computed over the evaluated portion of each clip.
MuscleMimic's training times were estimated from its reported training throughput~\cite{li2026towards}.

We fine-tuned specialist policies from the generalist checkpoint at epoch 1,080, each on a single reference clip, using longer backpropagation windows on an H200.
For the $360^\circ$ jumping twist, reference initialization times were sampled uniformly across the clip.
Training times and tracking errors were evaluated as described for individual-motion imitation.

\subsubsection*{Task-driven predictive learning}
We trained predictive locomotion policies using the reduced musculoskeletal model with 290 trunk and lower-limb muscles, with observations of the model's state and commanded forward speed.
Episodes were initialized from fixed walking and standing states with small random perturbations, without a motion-tracking objective.
Walking policies were trained from scratch with mirror symmetry, using 512 environments, a backpropagation window of 64 steps and 2,000 epochs.

The walking reward per environment step is
\begin{align}
 r_{\mathrm{walk}}={}&w_v v+w_g G-w_a c_a-w_m c_m-w_f\mathbf 1_{\mathrm{fall}}\nonumber\\
 &-w_z[h_0-h]_+^2-w_h c_h,\label{eq:walkreward}\\
 c_a={}&\frac{a_0^{2-p}}{N}\sum_{m=1}^N a_m^p,
 \qquad [x]_+=\max(x,0),
\end{align}
where $v$ is forward centre-of-mass (COM) speed, $G$ rewards speed tracking and movement stability, $c_a$ penalizes muscle activation with exponent $p$ across $N$ muscles, $c_m$ is the mass-normalized metabolic-rate estimate of Wang et al.~\cite{wang2012optimizing}, $\mathbf 1_{\mathrm{fall}}$ indicates termination for low COM height, and $h$ is COM height above the lowest point of the feet, penalized below $h_0$.
The tracking reward combines Gaussian factors for forward speed, lateral velocity, head angular velocity and head tilt,
\begin{equation}
 G=G_vG_yG_\omega G_b,\qquad
 G_v=\exp\!\left(-0.06\frac{[|v-v^*|-d_v]_+^2}{s_v^2}\right),
\end{equation}
where $v^*$ is the commanded speed, $d_v$ is the speed tolerance and $s_v$ is the scaling factor.
Head acceleration is penalized by
\begin{equation}
 c_h=\sum_{j\in\{x,z\}}\min\left([|\ddot p_{h,j}|/g_0-b_j]_+,2\right)^2,
\end{equation}
where $\ddot p_{h,j}$ is head acceleration along axis $j$, $g_0$ is gravitational acceleration and $b_j$ is the direction-specific tolerance.

We used an LLM agent to iteratively tune reward parameters based on quantitative gait measures from previous trials.
Each candidate policy was evaluated from eight initial states for sustained walking and agreement with human joint kinematics at approximately 1.3\,m/s~\cite{fukuchi2018public}.
We selected the final reward settings based on quantitative scores, gait plots and videos.
Human gait data thus informed reward selection but were not used as tracking targets during policy training.

Keeping the selected reward fixed, we trained policies at commanded speeds of 0.7--4.0\,m/s, both from scratch and by continuing training from lower-speed policies, with fixed or gradually increasing speed commands.
We evaluated the policies with sampled actions from eight initial states, analysing the final 5\,s of 10\,s rollouts.
Gaits were classified by the phase difference between COM potential and forward kinetic energies at the stride frequency: differences below $90^\circ$ indicate running-like mechanics, and larger differences indicate walking-like mechanics~\cite{falisse2019rapid}.
We further distinguished running with flight phases from running-like gaits without flight phases using an aerial-phase fraction threshold of 0.01.
COM vaulting during stance was quantified as COM height at mid-stance relative to the mean of its heights at foot strike and toe-off.

For faster predictive running, we modified the walking reward by removing the linear speed term and coupling the tracking reward with the muscle activation and metabolic costs,
\begin{equation}
 r_{\mathrm{gate}}=w_g G\exp[-\eta(w_a c_a+w_m c_m)]
 -w_f\mathbf 1_{\mathrm{fall}}-w_z[h_0-h]_+^2-w_hc_h,
 \label{eq:gatedreward}
\end{equation}
where $\eta=0.116$.
The representative running policy was trained by gradually increasing the commanded speed to 5.0\,m/s.

\subsubsection*{Human gait data and biomechanical evaluation}
We evaluated the agreement of our predicted walking gait with human data relative to four biologically inspired control models reported by Ton and Song~\cite{ton2026unified}.
These models represent distinct hypotheses of locomotor control: reflex-based sensory feedback~\cite{song2015neural}, central pattern generation combined with reflex feedback (CPG--reflex)~\cite{ogihara2001generation}, muscle synergies~\cite{aoi2019neuromusculoskeletal}, and a combination of all three mechanisms (CPG--reflex--synergy)~\cite{di2023investigating}, implemented in sagittal-plane musculoskeletal models with 18 muscles.
We used their simulation data and original root mean square error (RMSE) calculation procedure.
For each biomechanical variable, we linearly resampled the simulated curve to the number of samples in the corresponding human reference curve and computed the RMSE over the gait cycle.
We summed the individual RMSEs within three variable groups: kinematics, kinetics and muscle activation.
Kinematic errors are expressed in degrees, force and moment errors are normalized by body weight and body weight times leg length, respectively, and muscle activation errors are based on normalized activation envelopes.
Our simulated curves were averaged over 53 strides.
The human walking reference data comprise sagittal-plane trunk and joint angles from Rose and Gamble~\cite{rose2006human}, joint moments and ground reaction forces from Camargo et al.~\cite{camargo2021comprehensive}, and electromyographic envelopes from Perry and Burnfield~\cite{perry2010gait}.
For running, we compared our predicted gait with human treadmill running data at 3.5 and 4.5\,m/s~\cite{fukuchi2017running} and track running data at 5.17 and 6.96\,m/s~\cite{dorn2012muscular}.

Metabolic cost of transport (COT) was calculated after training using the metabolic model of Bhargava et al.~\cite{bhargava2004phenomenological}, as implemented by Falisse et al.~\cite{falisse2019rapid}, including a basal metabolic rate of 1.2\,W/kg.
COT was calculated as the metabolic energy expenditure per unit body mass and distance travelled.
Using the achieved locomotion speeds, we fitted a quadratic function to the lowest walking COT at each commanded speed and a linear function to the COT of all successful running and
running-like policies.
The minimum of the quadratic fit and the intersection of the two fitted curves estimate the energetically optimal walking speed and the energetic crossover speed, respectively.

\subsection*{Effects of tendon elasticity on predicted movement}
To investigate the effects of tendon elasticity, we trained policies on elastic- and rigid-tendon variants of the same musculoskeletal model with matched task-specific rewards and training settings.
The rigid-tendon variant uses Millard's inextensible-tendon formulation~\cite{millard2013flexing}, fixing tendon length at its slack length while retaining the muscle parameters.
Removing tendon compliance thus affects both elastic energy storage and fibre operating lengths under load.

For walking, we trained policies from scratch using the selected walking reward at a commanded speed of 1.3\,m/s and evaluated their gait biomechanics as described above.
For maximum-speed sprinting, we used a reward adapted from Bolt~\cite{wang2026bolt},
\begin{equation}
 r_{\mathrm{sprint}}=v_x+\exp(-2.5z^2)
 -0.03\sum_j|\tau_{\mathrm{lim},j}|,
 \label{eq:runreward}
\end{equation}
where $v_x$ is forward pelvis velocity, $z$ is lateral pelvis displacement from the lane centre and $\tau_{\mathrm{lim},j}$ is the joint-limit torque.
We trained two policies per tendon variant from different random seeds, each for 2,000 epochs with 512 environments and a backpropagation window of 64 steps.
Sprinting speed was evaluated over the final 5\,s of 10\,s rollouts from eight initial states.

For maximum-height jumping, we initialized policies from their respective motion-imitation policies trained on a recorded jump and continued training without motion tracking or effort penalties.
We rewarded increases in the running maximum of the predicted ballistic apex,
\begin{equation}
 A_t=z_c(t)+[\dot z_c(t)]_+^2/(2g_0),
\end{equation}
\begin{equation}
 r_t=s\left(\max_{0\le k\le t}A_k
 -\max_{0\le k<t}A_k\right),
\end{equation}
where $z_c$ and $\dot z_c$ are COM height and vertical velocity, respectively, and $s$ is a reward scale.
We conducted three independent training runs per tendon variant and selected the checkpoint with the greatest measured COM rise above standing height in each run, regardless of successful landing.

\section*{Data availability}
No experimental data were collected in this study.
The human motion data used for imitation learning are publicly available from the KIT Whole-Body Human Motion Database~\cite{mandery2015kit} and AMASS~\cite{mahmood2019amass}; we used the retargeted motions provided with MuscleMimic~\cite{li2026towards}.
The human gait and metabolic data used for comparison are available from the cited studies~\cite{rose2006human,camargo2021comprehensive,perry2010gait,fukuchi2017running,fukuchi2018public,dorn2012muscular,ralston1958energy,batliner2018metabolic}, and the simulation results of the four control models from Ton and Song~\cite{ton2026unified}.
The simulation data supporting the findings of this study
are available at
\url{https://github.com/neumovelab/diffneusim-experiments}.

\section*{Code availability}

DiffNeuSim's simulation engine and musculoskeletal models are available at
\url{https://github.com/neumovelab/diffneusim}.
The policy learning frameworks, training configurations, evaluation scripts, trained policies,
experimental results, and analysis code supporting this study
are available at
\url{https://github.com/neumovelab/diffneusim-experiments}.

\bibliographystyle{unsrtnat}
\bibliography{references,references_new}

\section*{Acknowledgements}
Research reported in this publication was supported in part by the National Institute on Aging of the National Institutes of Health under award number RF1AG096055.
The content is solely the responsibility of the authors and does not necessarily represent the official views of the National Institutes of Health.

\section*{Author contributions}
H.S. and S.S. conceived and designed the study.
H.S. developed the simulator, implemented the learning frameworks and conducted the computational experiments.
H.S. and S.S. analysed and interpreted the results.
S.S. drafted the manuscript with contributions from H.S\@.
S.S. supervised the project and secured funding.
Both authors reviewed and approved the final manuscript.

\section*{Competing interests}
The authors declare no competing interests.

\end{document}